\documentclass[runningheads]{llncs}

\usepackage[T1]{fontenc}

\usepackage{graphicx}
\usepackage{comment}
\usepackage{amsmath,amssymb}
\usepackage{color}
\usepackage{url}
\usepackage{hyperref}

\usepackage{times}
\usepackage{epsfig}
\usepackage{multirow}
\usepackage{array}
\usepackage[export]{adjustbox}

\newcolumntype{M}[1]{>{\centering\arraybackslash}m{#1}}
\newcommand{\specialcell}[2][c]{%
  \begin{tabular}[#1]{@{}c@{}}#2\end{tabular}}

\newif\ifreview
\reviewfalse

\ifreview
	\usepackage{lineno}

	\linenumbers
\fi

\begin{document}


\def\SubNumber{053}

\def\GCPRTrack{Main Track}

\title{Detection of Synthetic Face Images: Accuracy, Robustness, Generalization}

\ifreview
	\titlerunning{GCPR 2025 Submission \SubNumber{}. CONFIDENTIAL REVIEW COPY.}
	\authorrunning{GCPR 2025 Submission \SubNumber{}. CONFIDENTIAL REVIEW COPY.}
	\author{GCPR 2025 - \GCPRTrack{}}
	\institute{Paper ID \SubNumber}
\else

	\author{Nela Petr{\v z}elkov{\' a} \and
	Jan {\v C}ech \orcidID{0000-0002-2181-5917}}
	
	\authorrunning{N. Petr{\v z}elkov{\' a} and
	J. {\v C}ech}
	
	\institute{Faculty of Electrical Engineering, \\Czech Technical University in Prague, Czech Republic\\
	\email{cechj@fel.cvut.cz}}
\fi

\maketitle              

\begin{abstract}
An experimental study on synthetic face image detection is presented. We introduce FF5, a dataset of five fake face generators, including recent diffusion models. A baseline model trained on a specific generator achieves near-perfect accuracy in distinguishing synthetic from real images and handles common distortions (e.g., compression) via data augmentation. Additionally, partial manipulations, where synthetic content is blended into real images, can be detected and localized using a YOLO-based model.
However, the model is vulnerable to adversarial attacks and fails to generalize to unseen generators -- a limitation shared by state-of-the-art methods. Testing on Realistic Vision, a fine-tuned version of Stable Diffusion, confirms these challenges. Our study provides a quantitative evaluation of key properties and empirical evidence that deepfake detectors primarily learn generator fingerprints embedded in the signal.

\end{abstract}

\section{Introduction}

\begin{figure}
   \centering
   \includegraphics[width=0.76\linewidth]{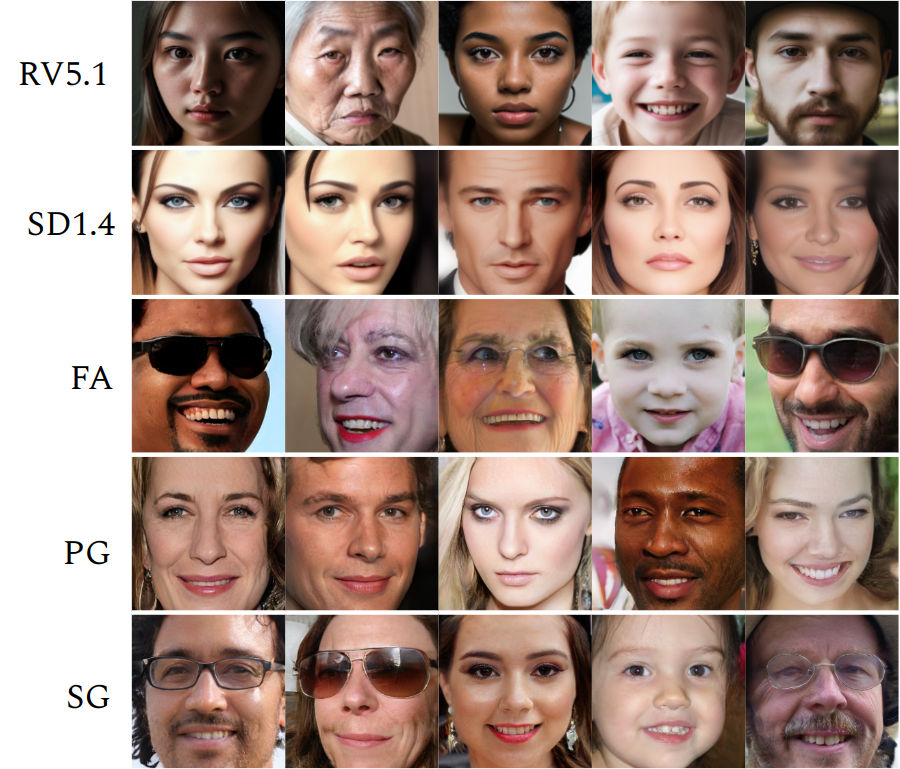}
   \caption{Samples of our FF5 dataset produced by five generators: two diffusion models -- Realistic Vision 5.1~\cite{RV} (RV5.1) and Stable Diffusion 1.4~\cite{Rombach-CVPR-2022_sd} (SD1.4), one commercial app -- FaceApp~\cite{FaceApp} (FA), two GANs -- Progressive GAN~\cite{PG-GAN} (PG) and StyleGAN~\cite{StyleGAN} (SG).}
  \label{fig:dataset}
\end{figure}

Image synthesis has made remarkable progress in recent years, thanks to the advances of generative models such as Generative Adversarial Networks (GANs)~\cite{Karras-CVPR-2020_stylegan2} and Diffusion Models~\cite{Rombach-CVPR-2022_sd}. Synthesized images are becoming increasingly realistic and hardly distinguishable from real ones to the naked eye of an average human and even of an expert, see Fig.~\ref{fig:dataset}. However, this progress also poses serious threats to individuals and society~\cite{Hancock-2021_social,Kietzmann-2020_social}, as synthesized images, also known as `deep fakes'~\cite{Nguyen-2022_deepfakes}, can be used for malicious purposes, such as fake porn~\cite{fakeporn}, fake video calls~\cite{fakecall,fakecall_klitchko}, fake news~\cite{zelensky_deepfake}, or fake videos in election campaigns~\cite{fake_elections_turkey,fake_elections_slovakia}. Therefore, it is important to develop effective and robust methods to detect and expose fake images, especially in the domain of faces, which are often the target of the attacks.

In this paper, we present a comprehensive experimental study that uncovers key properties of neural fake-face detectors. Rather than solely optimizing accuracy on standard datasets, we take a broader approach, using models with standard architectures to explore fundamental challenges in synthetic image detection. Specifically, we investigate the generalization ability of detectors when faced with unseen generators, their robustness to various image degradations and input sizes, their vulnerability to adversarial attacks, and their effectiveness in localizing manipulated regions within partially altered real images. 

In addition to detectors which are accurate in spotting recent generator images (namely, the Stable diffusion~\cite{Rombach-CVPR-2022_sd} –- Realistic Vision~\cite{RV}), our main contribution is a thorough analysis of forgery detectors that revealed many intriguing properties. 
To the best of our knowledge, no existing work offers a similar in-depth analysis within a compact and easily reproducible setup.
Our contributions are summarized below.

\begin{enumerate}
\item {\bf Novel FF5 dataset.} We collected a dataset of five fake face image generators. We extended DFFD corpus~\cite{DFFD-Dataset} by images produced by two recent diffusion model generators. 

\item {\bf Cross-generator-detector testing.}
We show that while it is surprisingly easy to train a detector for a specific synthetic image generator, its accuracy drops dramatically when tested on images produced by a different generator, which was not trained for. This effect is not much reduced by training the detector on images generated by multiple different generators. We quantify this effect and show learning curves that demonstrate the accuracy as a function of a number of training images.

\item {\bf Robustness to input size and degradations analysis.}
We tested the detector against blur (reflecting input resolution), JPEG compression, and input patch size reduction via masking. It demonstrated strong robustness, further improving when degradations were included as data augmentation. Notably, our detector spots synthetic images from a 25$\times$25 px patch with about 70\% accuracy.

\item {\bf Adversarial attacks vulnerability investigation.}
We demonstrate that adversarial images can easily be found to deceive the detector to classify synthetic image as real. Moreover, we show that residuals found for a particular detector model can also fool other models of a very different architecture. We tested both convolutional networks and a vision transformer. 

\item {\bf Localizing partial manipulations.}
A likely scenario of a fraudulent act is to blend synthetic images into real photos. Therefore, we prepared a set of partially manipulated images using state-of-the-art inpainting models replacing key regions of the face (eyes, nose, mouth, etc.).  We show that such images are easily spotted despite the manipulated area being small. Moreover, the manipulated area is localized within the image with high accuracy. 
\end{enumerate}

The rest of the paper is organized as follows. The related work is summarized in Sec.~\ref{sec:related_work}, the proposed methodology and results of the experimental analysis are presented in Sec.~\ref{sec:experiments}, and finally, Sec.~\ref{sec:conclusion} gives conclusions.

\section{Related Work} \label{sec:related_work}

In conjunction with the rapid development of high-quality synthetic image generators, research on the detection of fake images has become very active. For a comprehensive review, we refer to recent surveys~\cite{liu2025review_survey,DF40,heidari2024deepfake_survey} or a handbook~\cite{rathgeb2022handbook}.   
In this section, we review some of the existing methods and challenges for this problem.

Historically, before the boom of deep learning, fake image detection focused on detecting ``doctored images'' that were manually edited or manipulated from images captured by cameras. These methods relied on various clues, such as steganographic features, compression artifacts, or inconsistencies in lighting or shadows~\cite{Farid-2009,Rocha-2011}. However, these methods are not effective against synthetic images that are generated from scratch or with minimal human intervention.

Forensic low-level signal detectors are another class of methods. They exploit the spectral signatures of synthetic images. Inspiration probably came from the recognition of a camera device~\cite{Chen-2008}. More recently, researchers discovered that the residual spectra of synthetic images contain typical anomalies, which creates a spectral fingerprint of a synthetic generator~\cite{Efros-CVPR-2020,Corvi-ICASSP-2023,Corvi-CVPR-2023}.  A frequency domain method is presented in~\cite{Durall-2020}.  

In a similar spirit, other methods suggested that the information for fake image detection is deeply embedded in the image signals and can be detected independently of the image content. Chai et al.~\cite{Chai-ECCV-2020} used a CNN with a narrow receptive field to detect fake images from signal patches, highlighting hair as a key discriminative region. Tan et al.~\cite{Tan-CVPR-2024} proposed to spot upsampling artifacts by modeling neighboring pixel relationships. Shiohara et al.~\cite{Blending} proposed detecting blending artifacts in deepfake images.


Wang et al.~\cite{Efros-CVPR-2020} showed that GAN-generated images have distinctive features, making detection easy with a CNN classifier. Their model generalized well to unseen generators but was tested only on GAN-based images. Very promising approach to detect AI-generated images is by using CLIP~\cite{CLIP}, as a powerful image encoder, followed by a lightweight classifier head~\cite{Ojha-CVPR-2023,Cozzolino-CVPRw-2024,Yermakov-2025}. The authors report promising generalization abilities. 

In this paper, we demonstrate that, in the leave-one-out setup, the generalization of detectors to unseen generators is poor. We show that even recent state-of-the-art synthetic image detectors either fail completely or perform low when tested on images produced by novel unseen generators. A recent work related to ours~\cite{Park-CVPR-2025} tested generalization by learning on samples from large scale dataset comprising 4.8k distinct synthetic generator models. 

Besides detection, some recent works have also addressed the localization of fake images, which aims to segment the manipulated areas of the real images. The problem is challenging considering possibly a small area of manipulation. Some methods do not use any special architecture for localization, but rely on post-processing techniques. Recent paper~\cite{Tantaru-2024} compares a popular Grad-CAM~\cite{gradcam} to highlight the regions that contribute to the classification decision and the scanning technique of~\cite{Chai-ECCV-2020} to localize synthetic regions in partially manipulated images.  Other methods use more complex architectures, such as multi-branch network~\cite{Guo-CVPR-2023}, or dense self-attention network~\cite{Hao-ICCV-2021_transforensics}, to explicitly learn the localization maps. Paper~\cite{SAM_forgery} fine-tunes a large segmentation model (SAM)~\cite{sam} to adapt it to the fake image domain. We show that precise localization results are achieved for relatively small regions using a simple YOLO-based architecture~\cite{yolo},  namely YOLOv8~\cite{yolov8}, as long as the fake images are composed of images produced by the same generator model that was trained on.

Deep neural networks are known to be vulnerable to adversarial attacks~\cite{Szegedy-ICLR-2014_adversarial}, which are small imperceptible perturbations of the input that cause a network to make a wrong prediction. This problem has been extensively studied in various domains, such as image classification~\cite{Goodfellow-ICLR-2015}, object detection~\cite{Wei-IJCAI-2019}, or face recognition~\cite{Dong-CVPR-2019}. In this paper, we show that this vulnerability also applies to the fake image detection domain and that a common way of generating adversarial examples can fool the detectors into classifying fake images as real.



\section{Methodology and Experimental Results} \label{sec:experiments}


%
Our FF5 dataset consists of face images produced by five generators; see Fig.~\ref{fig:dataset}. We use two diffusion models: Realistic Vision V5.1~\cite{RV} (RV5.1), which is fine-tuned Stable Diffusion sharing the same architecture, and official StabilityAI's Stable Diffusion V1.4~\cite{Rombach-CVPR-2022_sd} (SD1.4). Then three synthetic sets that are part of the DFFD corpus~\cite{DFFD-Dataset}: FaceApp~\cite{FaceApp} (FA), which are images produced by a popular commercial mobile phone application with undisclosed technology, and GANs PG-GAN2~\cite{PG-GAN} (PG), and StyleGAN~\cite{StyleGAN} (SG). 

For the diffusion models, we used dynamic prompting~\cite{dynamic_prompt} which enables us to automatically alter a prompt with terms from predefined options. Our base prompt was ``RAW photo'' and we randomly altered it with attributes influencing the gender, age, accessories, and the environment. See Sec.~\ref{sec:implementace} for more details. 
That enabled us to quickly generate diverse images. With different random seeds, we generated almost 1.7k images for each diffusion model. The other generators consist of 2k images for each of FA, PG, and SG. 

For the negative class of real images, we use images from the FFHQ dataset~\cite{StyleGAN}. All synthetic and real images underwent the same preprocessing procedure, aligning using facial landmarks, cropping with the same margin, and resampling to $224 \times 224$~px. 

\subsection{Cross-generator testing} \label{sec:cross-generator}

\begin{table}
    \normalsize
    \centering
    \caption{Cross-generator testing.  Each cell $(row,col)$ shows test accuracy in percent of the models trained (a) on generator $row$ / (b) all without generator $row$, tested on generator $col$.}
    \label{tab:one_source_only_classification}    
    \begin{tabular}{|M{0.3cm}|c|c|c|c|c|c|}
    \cline{3-7}
     \multicolumn{2}{c|}{} & \multicolumn{5}{c|}{Test set} \\ \cline{3-7}
     \multicolumn{2}{c|}{} & RV5.1 & SD1.4 & FA & PG & SG \\  \hline
     \multirow{5}{*}{\rotatebox{90}{Training set~}} & RV5.1 & \textbf{100} & 58 & 49 & 50 & 50 \\ \cline{2-7}
     & SD1.4 & 51 & \textbf{100} & 50 & 54 & 49 \\ \cline{2-7}
     & FA & 53 & 50 & \textbf{80} & 87 & 60   \\ \cline{2-7}
     & PG & 49 & 61 & 54 & \textbf{100} & 50  \\ \cline{2-7}
     & SG & 48 & 48 & 54 & 66 & \textbf{94}  \\ \hline
    \end{tabular}
    \hfill
    \begin{tabular}{|M{0.25cm}|c|c|c|c|c|c|}
    \cline{3-7}
    \multicolumn{2}{c|}{} & \multicolumn{5}{c|}{Test set} \\\cline{3-7}
     \multicolumn{2}{c|}{}  & RV5.1 & SD1.4 & FA & PG & SG \\  \hline
     \multirow{5}{*}{\rotatebox{90}{Training set~}} & -RV5.1 & \textbf{58} & 92 & 80 & 94 & 91  \\ \cline{2-7}
     & -SD1.4 & 91 & \textbf{84} & 85 & 91 & 91  \\ \cline{2-7}
     & -FA & 95 & 94 & \textbf{55} & 94 & 89  \\ \cline{2-7}
     & -PG & 93 & 92 & 77 & \textbf{79} & 85  \\ \cline{2-7}
     & -SG & 95 & 95 & 80 & 94 & \textbf{52}  \\ \hline
    \end{tabular}\\[1ex]
      ~~ (a) Training on a single generator \hfill (b) Leave-one-out training ~~~~~~

\end{table}

In this experiment, we trained the ResNET-50 backbone binary classifiers~\cite{Resnet} between synthetic and real samples. This is the same architecture used by~\cite{Efros-CVPR-2020}. The dataset was always split to 80-10-10\% for disjoint training-validation-test sets, respectively. The ratio between synthetic and real classes was always 50-50\%. We used Adam optimizer with default settings and horizontal flipping as data augmentation. We always selected the model that achieved the best accuracy on the validation set.

We performed the following cross-generator experiment. We first trained on single-generator images and tested on all in the set, see Tab.~\ref{tab:one_source_only_classification}a. Then, the other way around, we trained on all generators with one left out and tested again on all, see Tab.~\ref{tab:one_source_only_classification}b.

%

We can see in Tab.~\ref{tab:one_source_only_classification}a that if the detector is trained on the same model as it is tested (diagonal of the table), the accuracy is perfect for RV5.1, SD 1.4, PG, and very high for SG. The accuracy is only 80\% for FA. FA, FaceApp~\cite{FaceApp}, a commercial app with unknown technology behind, probably blends the real face with some manipulations, making it harder to identify. However, we can clearly see (off the diagonal) that accuracy drops close to chance when we test on images produced by models for which the detector was not trained on. Interestingly, this is not the case of FA, which achieves even higher accuracy on PG, which might indicate similar technology, but the converse is not true. The generalization does not occur for even very similar models, the diffusion models RV5.1 and SD 1.4 share the same architecture. 

In Tab.~\ref{tab:one_source_only_classification}b, when the detector is trained on multiple models, a certain level of generalization to unseen generators is achieved for some models, as seen in the diagonal now. SD1.4 seems to generalize well while it was not trained on it. However, RV5.1 is fine-tuned version of SD1.4, but the generalization is not reciprocal. PG seems to generalize partially as it is another GAN as SG. The rest is close to chance. 


Note that the cross-dataset experiment includes several generators representing the fake class, while the real class is represented by a single source, the FFHQ dataset. This is a limitation, as real-world face images exhibit significantly more diversity than what is captured by this dataset. Therefore, these results should be interpreted as an optimistic upper bound; accuracy is expected to decline when a more diverse and previously unseen real dataset is used.

\bigskip

\begin{figure}[t]
    \centering
    \includegraphics[width=0.8\columnwidth]{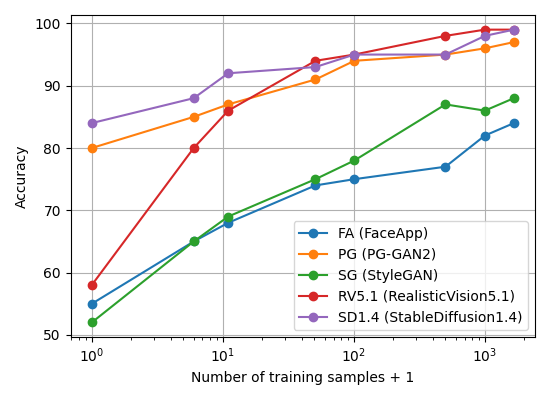}
    \caption{Learning curves for training a detector to spot images produced by a new generator. Test accuracy as a function of the number of training samples. Horizontal axis is logarithmic.}
    \label{fig:num_samples_in_train_set}
\end{figure}

We see that the detector generalization to an unseen model is a problem. Therefore, in the following experiment, we measure how many samples of the new generator are needed for fine-tuning. We always start from the model that is trained on all the generators of our set except one (i.e., the rows of Tab.~\ref{tab:one_source_only_classification}b), so its initial accuracy is on the diagonal of Tab.~\ref{tab:one_source_only_classification}b. Then we gradually add training samples of the new model (0, 5, 10, 50, 100, 500, 1000, 1666) samples and measure the accuracy on the test set. The results are shown as learning curves in Fig.~\ref{fig:num_samples_in_train_set}. Note that the plot has a logarithmic horizontal axis. 

Interestingly, the learning curves are steep. For some generators, only a few units or small tens of training samples are sufficient to significantly improve detection accuracy, indicating that the model quickly captures the fingerprint of the new generator.

\paragraph*{Comparison with the state of the art.} We evaluate recent fake image detectors on our test set produced by the RV5.1 generator, see Tab.~\ref{tab:sota}. The first four methods~\cite{Efros-CVPR-2020,Guo-CVPR-2023,Tan-CVPR-2024,Cozzolino-CVPRw-2024} provide pre-trained models, while the last two models, Durall~\cite{Durall-2020} and our ResNET-50, were trained on an independent training split of the RV5.1 dataset.

Wang~\cite{Efros-CVPR-2020} claims to generalize to unseen generators, but this does not hold for novel diffusion models such as RV5.1. The model achieved accuracy close to chance, likely because it was trained on GAN-based generators, which do not generalize to the recent diffusion-based RV5.1 generator. HiFi~\cite{Guo-CVPR-2023} failed despite being trained on diffusion models. Tan~\cite{Tan-CVPR-2024} performed slightly above chance level, even though the paper reports a generalization to unseen generators by spotting upsampling artifacts. Cozzolino~\cite{Cozzolino-CVPRw-2024} achieves better, but still low, accuracy despite being trained on Stable Diffusion and reporting generalization abilities via CLIP~\cite{CLIP}.

\begin{table}
   \normalsize
   \caption{Comparison with the state of the art. Accuracy on test set produced by RV5.1 generator. Last two models were trained on independent split of RV5.1 dataset.}   
   \centering
   \tabcolsep=1.9pt
    \begin{tabular}{|c||c|c|c|c||c|c|}  \hline
     Model & Wang~\cite{Efros-CVPR-2020} & HiFi~\cite{Guo-CVPR-2023} & Tan~\cite{Tan-CVPR-2024} & Cozzolino~\cite{Cozzolino-CVPRw-2024} & Durall~\cite{Durall-2020} & ResNET-50\\  
     ~ & CVPR'20 & CVPR'23 & CVPR'24 &  CVPRw'24 & CVPR'20 & ours \\ \hline
     Accuracy &  48.3 &  44.2 & 64.1 & 70.0 & 87.7 & 99.5  \\     \hline
    \end{tabular}
    \label{tab:sota}
\end{table}

Our simple ResNET-50, when trained on the RV5.1 training split, achieved near-perfect recognition. In contrast, Durall~\cite{Durall-2020} resulted in inferior accuracy, likely due to its reliance on very simple features -- magnitude spectrum radius and logistic regression.

This experiment demonstrates that generalization to unseen generators remains an unsolved problem in practice. A trivial classifier, when trained on examples from the target generator, outperforms more sophisticated methods. The likely reason is that detectors overfit to known generator fingerprint and are unable to identify more universal traces that separate synthetic and real samples. 

The remaining experiments are conducted using our models trained within the dataset, since the competing methods do not generalize well and some perform at chance level, as seen in Tab.~\ref{tab:sota}.

\subsection{Detector accuracy for input degradation} \label{sec:degradations}

\begin{figure}
\def\wdth{0.5925\linewidth}
\centering
\includegraphics[width=\wdth]{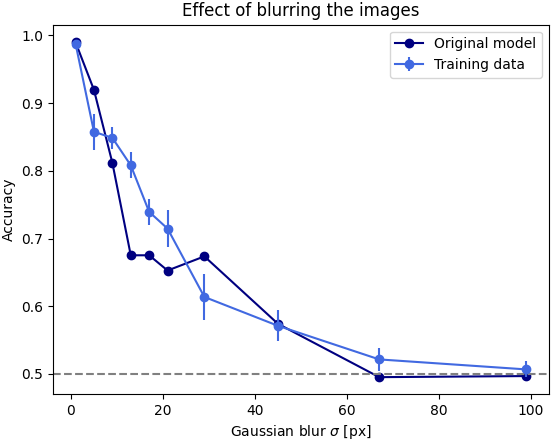}
\includegraphics[width=\wdth]{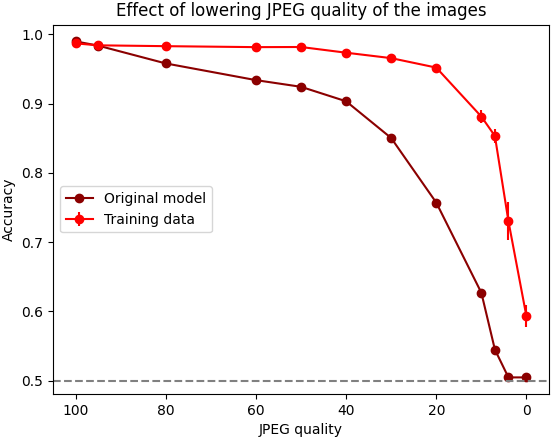}
\includegraphics[width=\wdth]{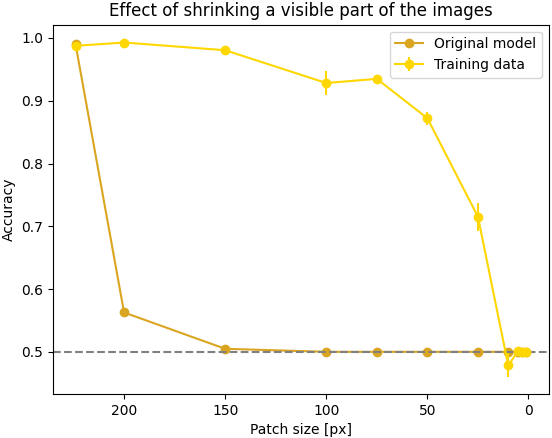}
\caption{Detector accuracy for input degradation. From top to bottom: Gaussian blur, JPEG compression, input patch size. Two scenarios are tested: (1) the detector is trained on undistorted images only, (2) the detector is trained on images including the degradations. Plots have error bars of standard deviation across 10 training trials.}
\label{fig:degradation}
\end{figure}

\begin{figure}
\centering
\includegraphics[width=1\linewidth]{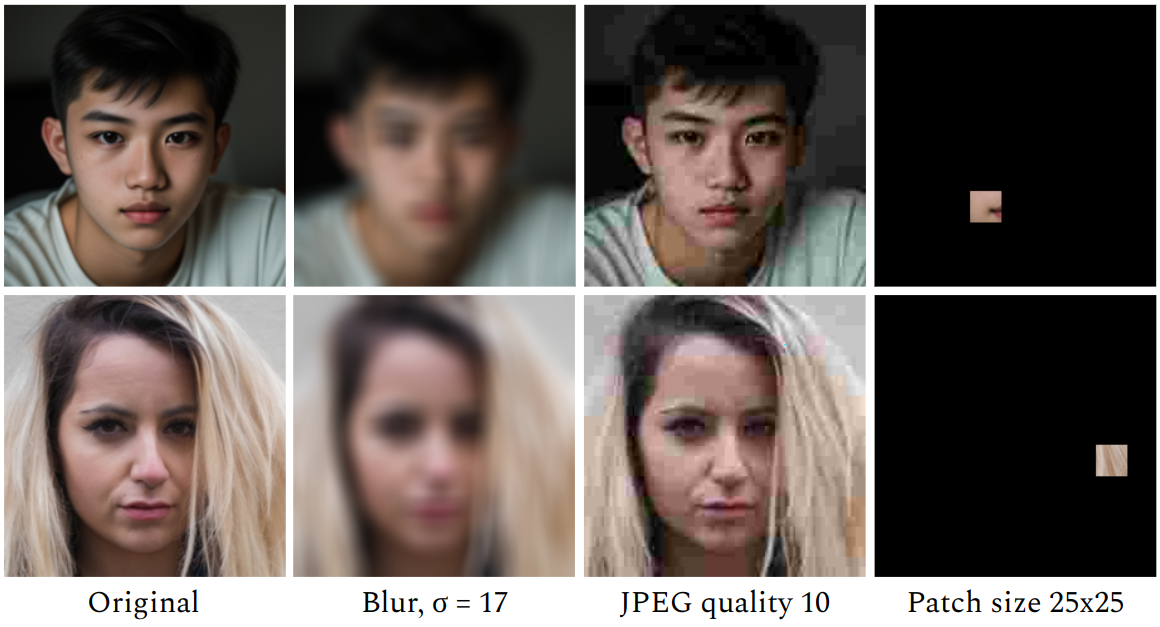}
\caption{Examples of distorted images.}
\label{fig:distorted_images}
\end{figure}


Since the image may be distorted, e.g., resized, compressed, cropped, prior to the distribution, we measured detector accuracy for the distortion. Gaussian blur simulates shrinking the resolution, and JPEG a lossy compression. The size of the input was simulated by masking the input image -- a square patch of a given size at random position is kept, while all other pixels are replaced with zeros in all three RGB channels. See Fig.~\ref{fig:distorted_images} for some examples.

We evaluated two scenarios. First, we tested the original model, which was trained on undistorted images from the RV5.1 set. Second, we re-trained the detector with the image degradation as data augmentation. 

The results are shown in Fig.~\ref{fig:degradation}. We can see that the detector proves a good robustness to the degradations, especially for the second scenario with re-training. For instance, the detector achieves accuracy about 80\% for Gaussian blur $\sigma = 17$ px, 90\% for JPEG quality 10, and 70\% for a patch as small as 25$\times$25 px.  

These findings corroborate that the fingerprint is strong, survives severe image degradations, and can be identified through discriminative learning. The small-patch experiment proves that the fingerprint exists at a low (signal) rather than a high (semantic) level.





\subsection{Adversarial attacks}



In this section, we study the vulnerability of our detector to adversarial attacks. An adversarial attack means performing a hardly perceptible modification of an image (residue) that causes a change in classification of the detector, i.e., a synthetic image is classified as real. First, we find the adversarial residua for a given image. 
We use the fast gradient sign method (FGSM)~\cite{Goodfellow-ICLR-2015}. The adversarial residuum is then an image of the same size as the input receptive field of the model that contains signs of the gradient of the output class score with respect to the input pixel intensities in each channel. This is easily calculated by backpropagation without any optimization.  

We tested the attacks on three different architectures of detectors: ResNET-50~\cite{Resnet}, Xception~\cite{xception}, ViT-tiny~\cite{vit-tiny}, 
which we trained on our training set RV5.1. All models achieved a perfect 100\% accuracy on the test set. 

\begin{figure}[t]
    \normalsize
    \bigskip
    \setlength\tabcolsep{4.5pt}
    \centering
    \parbox[c]{0.5\linewidth}{%
    \raggedright
    \includegraphics[width=0.9\linewidth]{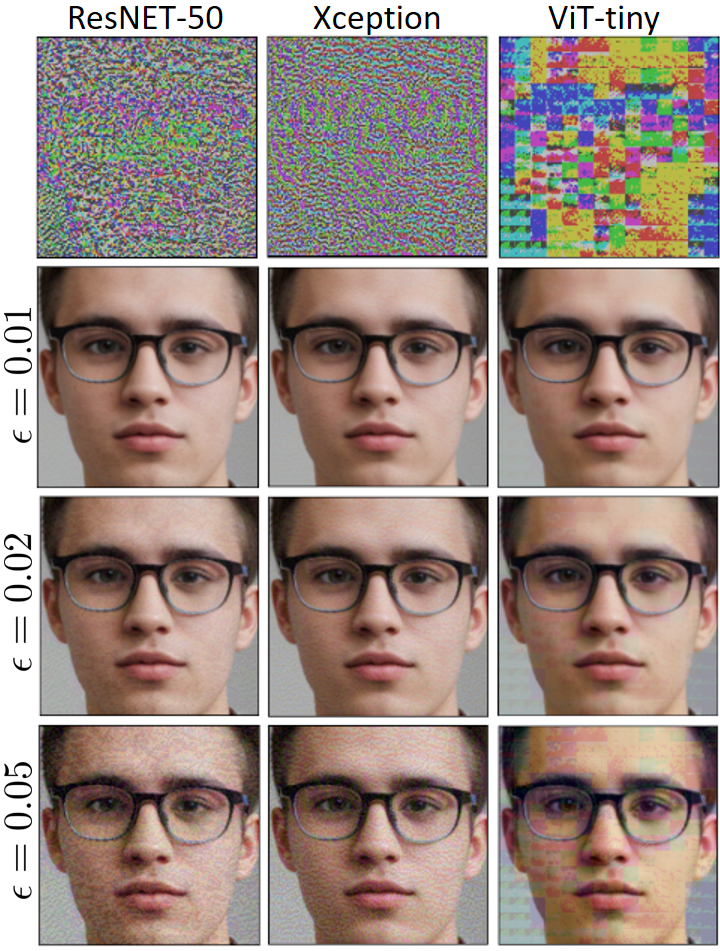}}%
    \parbox[c]{0.5\linewidth}{%
    \tabcolsep=1pt
     \begin{tabular}{|c|c|c|c|}
    \hline
    \textbf{\specialcell{FGSM\\ $\epsilon = 0.01$ }} & ResNET-50 & Xception & ViT-tiny \\ \hline
    ResNET-50 & \textbf{100.00} & 67.15 & 7.30 \\ \hline
    Xception & 2.55 & \textbf{100.00} & 6.93 \\ \hline
    ViT-tiny & 0.39 & 6.64 & \textbf{59.77} \\ \hline
    \end{tabular}
    ~\\[1.5ex]
    \begin{tabular}{|c|c|c|c|}
    \hline
    \textbf{\specialcell{FGSM\\ $\epsilon = 0.02$}} & ResNET-50 & Xception & ViT-tiny \\ \hline
    ResNET-50 & \textbf{100.00} & 100.00 & 8.76 \\ \hline
    Xception & 48.54 & \textbf{100.00} & 8.39 \\ \hline
    ViT-tiny & 16.80 & 89.06 & \textbf{92.97} \\ \hline
    \end{tabular}
    ~\\[1.5ex]
    \begin{tabular}{|c|c|c|c|}
    \hline
    \textbf{\specialcell{FGSM\\ $\epsilon = 0.05$}} & ResNET-50 & Xception & ViT-tiny \\ \hline
    ResNET-50 & \textbf{100.00} & 100.0 & 15.69 \\ \hline
    Xception & 100.00 & \textbf{100.0} & 10.22 \\ \hline
    ViT-tiny & 100.00 & 100.0 & \textbf{100.0} \\ \hline
    \end{tabular}}%
         \caption{Cross-architecture adversarial attacks for increasing strength of the residua $\epsilon$. Left: Examples of adversarial residua for specific models created with FGSM method. Right: Results showing confusion rate in percent for each cell $(row, col)$. The attack was targeted against model of architecture in $row$ and tested against the model of architecture in $col$.}    
    \label{tab:adversarial_attack_cross_model}
\end{figure}

The resulting residua are shown in 
Fig.~\ref{tab:adversarial_attack_cross_model}.
The residua are scaled up multiple times in the figures, otherwise the pattern would not be visible. All these residua scaled by strength $\epsilon$ if summed with the original images will switch the classification of the corresponding detector to ``real". Residua for different models appear different and the pattern is visibly influenced by the structure of the original image, as seen in Fig.~\ref{tab:adversarial_attack_cross_model}.


We measured the success of attacks by the \emph{confusion rate}, which depicts a percentage of test cases when the model switched the classification due to the attack from ``fake'' to ``real'' over the number of ``fake'' decisions prior to the attack. 


We tested the cross-model scenario, where the adversarial residua are found for a given image and a given model, and are tested also on other models of different architectures. The results for increasing strengths of the residua, $\epsilon$, are summarized in the tables in Fig.~\ref{tab:adversarial_attack_cross_model}.  

We see that for a small strength $\epsilon=0.01$, in-model attacks (in the diagonal) are successful for ResNET-50 and Xception. However, for higher strength, cross-model attacks, where a model of \emph{different architecture unknown by the attacker} works also, as seen off the diagonal. For strength $\epsilon=0.05$ a residue found for ViT-tiny confuses ResNET-50 and Xception too. 

A defense against adversarial attacks on deepfake detectors should be implemented in practice. It has been reported, e.g., in~\cite{Mumcu-2024}, that image compression or low-pass filtering can mitigate adversarial effects. However, as we have shown in this paper, such techniques also weaken the deepfake generator fingerprint. Another option is to use adversarial training~\cite{zhao2024adversarial} or detectors for adversarial patterns, though this remains a non-trivial problem~\cite{tramer2022detecting}, similar in spirit to deepfake detection. Therefore, we believe these problems should be studied together in future research.

\subsection{Localizing partial manipulations}

A likely scenario for constructing a fake image is that a synthetic image is seamlessly blended into a real face image. In this experiment, we will show that these partial manipulations are easy to identify together with localizing the area of the manipulations. 

We first prepared a dataset of partially manipulated face images. We randomly sampled real faces (from the FFHQ dataset) and, for each image, uniformly changed either of the eyes, eyebrows, nose, or mouth. These regions were detected using facial landmarks~\cite{FL-dlib09}, and the change of content within the region was carried out by Stable Diffusion inpainting~\cite{Rombach-CVPR-2022_sd}. This way we produced a dataset of 3.2k partially manipulated images that were mixed with 540 real images. 

The data set was divided into training, validation, and test subsets with proportions of 80\%, 16\%, and 4\%, respectively. Then, we trained YOLOv8~\cite{yolov8}, which is a YOLO-based architecture~\cite{yolo} with a segmentation head. 

\begin{figure*}[t]
   \centering
   \includegraphics[width=1\linewidth]{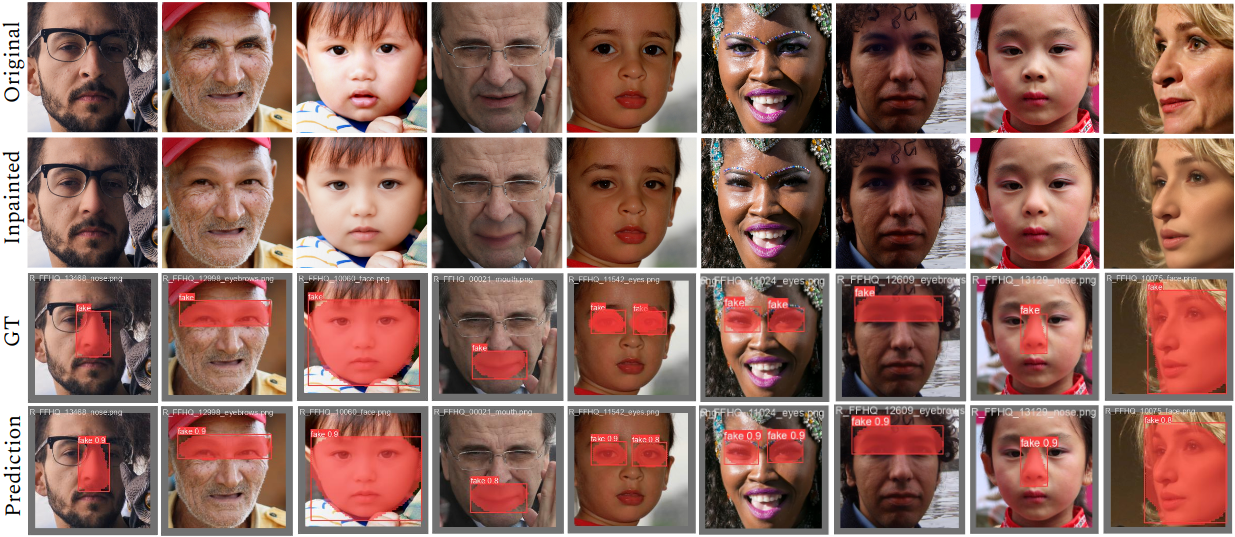}
   \caption{Localizing partial image manipulations perpetrated by inpainting of the ground-truth (GT) regions for examples of the test set. Localization predictions were found by our YOLOv8-based model.}
   \label{fig:localization_predictions}
\end{figure*}

Qualitative results on the test set are shown in Fig.~\ref{fig:localization_predictions}. It is seen that detected regions are found precisely, despite the fact that the manipulated (synthetic region) is sometimes fairly small with respect to the entire (real) image and no obvious artifacts are visible in the images. 

Quantitatively, the detector achieved mAP50 98\% (mean average precision for 50\% prediction/ground-truth detection overlap by intersection over the union). Pixelwise recall and precision were $95\%$ and $91\%$, respectively. 

We compared the detector with HiFi~\cite{Guo-CVPR-2023} which is supposed to provide localization of the manipulation. However, this detector failed completely and always recognized all our partially manipulated images as real. This again confirms, similarly to our findings in Sec.~\ref{sec:cross-generator}, that generalization to localize partial manipulations when using unseen models is very challenging. 

On the other hand, we attribute the success of the detector trained for this particular manipulation technique to its sensitivity in identifying small patches of the synthetic signal, as discussed in Sec.~\ref{sec:degradations}, and possibly to its ability to detect subtle boundary artifacts. The first reason is likely stronger, as the localization accuracy depends on the size of the manipulated area, which is quantified below.

\begin{figure}[t]
    \normalsize
    \centering
    \parbox[c]{0.5\linewidth}{\includegraphics[width=0.5\columnwidth]{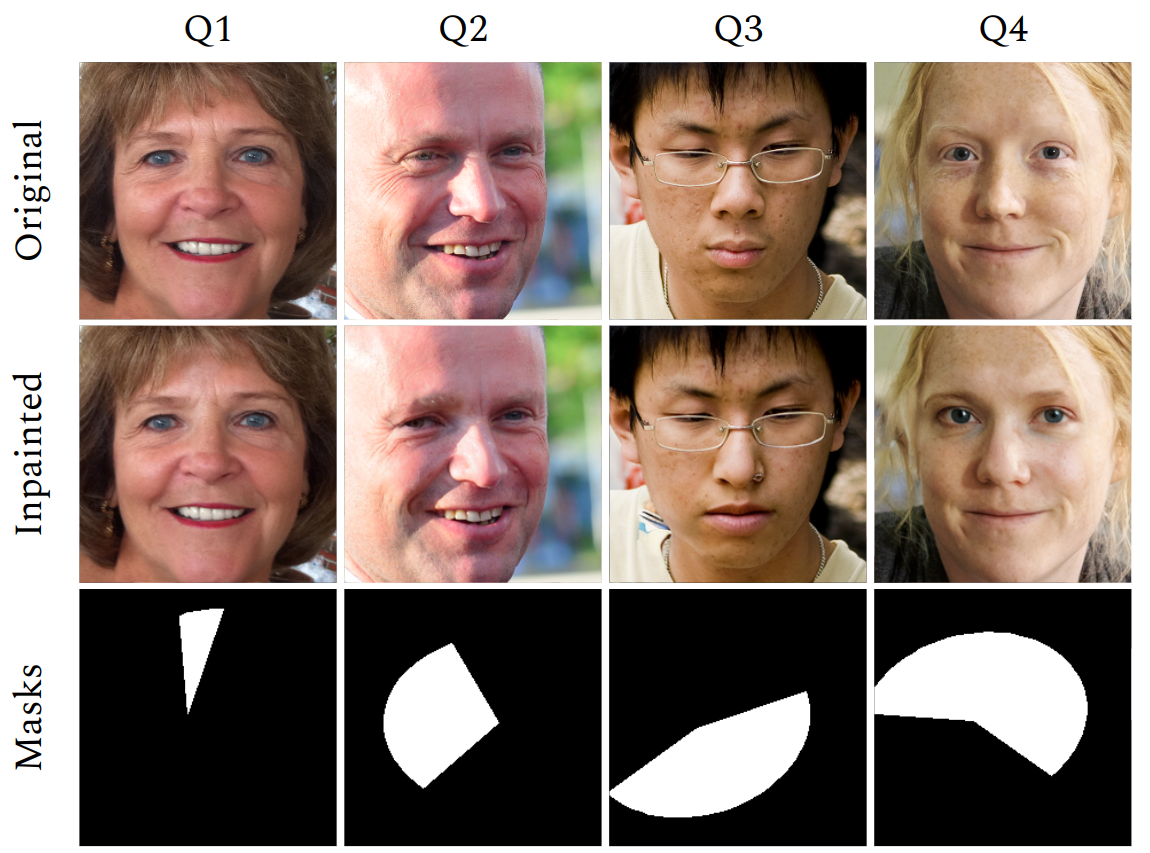}\\
    \small (a) Examples of partial image manipulations by inpainting with random masks. 
    \vfill
    }\hfill
    \parbox[c]{0.47\linewidth}{%
    \hfil\begin{tabular}{|c|c|c|c|c|} \hline
        & Q1 & Q2  & Q3 & Q4 \\ \hline
        Area size [\%] & 0--8 & 8--16 & 16--25 & 25--36 \\\hline
        mAP50 [\%] & 24 & 74 & 90 & 97 \\\hline
    \end{tabular}\hfill\\[1ex]
    \small{(b) Results -- mAP50 on the test images. Area size is the percentage of the manipulated area over the entire area of the face image.}
    \vfill
    }
    \caption{Localization accuracy as a function of manipulated area size.}    \label{fig:localization_size}
\end{figure}

\medskip
\paragraph*{Localization accuracy as a function of manipulated area size.} 
In this experiment, we quantify how the size of the manipulated area impacts the localization accuracy of our YOLOv8s-seg segmentation model. 
We created a dataset consisting of 4.5k of partially manipulated images with various sizes of the manipulated area. In particular, we randomly sampled real (FFHQ dataset) images, then for each image, we generated the manipulated area by a randomly placed, rotated, and cropped ellipse of random size. Finally, as in the previous experiment, we used Stable Diffusion's inpainting to modify the images in these areas. Several examples can be seen in Fig.~\ref{fig:localization_size}a. The model was trained on 2.7k of the 4.5k images, 1.8k were used for validation and testing.

The test set was split into equally sized bins Q1--Q4 according to the size of the manipulated areas: {'Q1': (0, 0.08), 'Q2': (0.08, 0.16), 'Q3': (0.16, 0.25), 'Q4': (0.25, 0.36)}, where the numbers denote a ratio of the area of the generated parts with respect to the total area. The results are shown in Fig.~\ref{fig:localization_size}b, where mAP50 is evaluated for each test set bin. We can see that it is obviously easier to localize larger areas for the model. Unlike in previous experiments, where we modified facial features (face, eyes, eyebrows, nose, mouth), here we chose the unpainted regions completely randomly. It happens that especially small regions are located in flat areas without texture. These regions do not manifest much of a usable signal for identification as do larger areas. This can be the reason why the smaller modified areas are more challenging to localize by the model.

\subsection{Implementation details} \label{sec:implementace}

We used AUTOMATIC1111's Stable Diffusion Web UI~\cite{webui} for our experiments. Besides graphical interface with many plugins, it also provides a convenient batch processing. 

To generate our dataset (RV5.1 and SD1.4), we used dynamic prompt~\cite{dynamic_prompt}. This is the Web UI extension that implements an expressive template language for the generation of random or combinatorial prompts. In particular, we used the following prompt to get diversity in our datasat: ``RAW photo, \{older $|$ younger\} \{man $|$ woman $|$ lady $|$ girl $|$ boy\} \{ \{smiling $|$ staring\} $|$ with glasses $|$ with hat $|$ with \{brown $|$ blonde $|$ dark\} \{straight $|$ curly $|$ short\} hair \}, high quality portrait taken with Nikon camera, in \{nature $|$ a city $|$ a room $|$ an office $|$ a park $|$ a street $|$ a forest \}''.

All models were trained with PyTorch framework.


\section{Conclusion} \label{sec:conclusion}

In this paper, we conducted several experiments on the detection of synthetic face images. Our results allow us to draw the following conclusions.

The \emph{good news} is that it is possible (if the generator of synthetic images is available) to train a simple model with an off-the-shelf architecture, which has almost perfect accuracy in distinguishing between synthetic and real images. The accuracy achieved far outperforms human abilities~\cite{Somoray-2023}. Another positive aspect is that the detector can be trained with data augmentation, to make it robust to common image distortions (reduced resolution, compression), and it can achieve good accuracy with only a small input patch from the face. Moreover, it is easy to detect the case of partial manipulations, where a collage of real and synthetic images is made. The manipulated area is automatically localized by training a standard model~\cite{yolov8}.

However, there are also \emph{bad news}. It is simple to prepare an adversarial attack. 
It turns out that the residua found for a target model act adversarially, even on other models of very different architectures. We showed that adversarial images found for vision transformers often confuse convolutional networks. 
The worst news is that the detectors do not generalize well to generators they were not trained on. This is not just the case of our simple detector, but we showed that many tested state-of-the-art detectors could not reliably detect synthetic images generated by a newer generator, which they were not trained on.

This study presents multiple insights from targeted experiments and provides quantitative evidence on the challenges of synthetic image detection. In summary, current detectors trained discriminatively in a supervised manner learn to identify signal fingerprints of specific generators. Future research will focus on novel learning strategies to mitigate such overfitting and enhance generalization.


\bigskip
\noindent
{\bf Acknowledgement.} This work was supported by the national recovery plan projects CEDMO NPO (1.4 CEDMO 1 -- Z220312000000) and CEDMO 2.0 NPO (MPO 60273/ 24/21300/21000) provided by the ministry of industry and trade, and by the CTU student grant SGS23/173/OHK3/3T/13. We thank to Patricie Petri{\v l}\!{\'a}kov{\'a} for evaluating the recent detectors on our dataset.

%
%
%
%
\bibliographystyle{splncs04}
\bibliography{053-main.bib}

\end{document}